%% file: main.tex
\documentclass[letterpaper]{article} 
\usepackage{aaai2026}  
\usepackage{times}  
\usepackage{helvet}  
\usepackage{courier}  
\usepackage[hyphens]{url}  
\usepackage{graphicx} 
\usepackage{natbib}  
\usepackage{caption} 
\usepackage{algorithm}
\usepackage{algorithmic}
\usepackage{amsmath}
\usepackage{amssymb}
\usepackage{booktabs}
\usepackage[table]{xcolor}
\usepackage{xcolor}
\usepackage{makecell}
\usepackage{multirow}
\usepackage{subcaption,subfig}
\usepackage{cleveref}
\usepackage{csquotes}
\usepackage{paralist}

\usepackage{newfloat}
\usepackage{listings}
\DeclareCaptionStyle{ruled}{labelfont=normalfont,labelsep=colon,strut=off} 
\floatstyle{ruled}
\newfloat{listing}{tb}{lst}{}
\floatname{listing}{Listing}
\title{Learning Spatial Decay for Vision Transformers}
\author{
    Yuxin Mao\textsuperscript{\rm 1}, Zhen Qin\textsuperscript{\rm 2}, Jinxing Zhou\textsuperscript{\rm 3}, Bin Fan\textsuperscript{\rm 1}, Jing Zhang\textsuperscript{\rm 1}, Yiran Zhong\textsuperscript{\rm 3}, Yuchao Dai\textsuperscript{\rm 1}\thanks{Corresponding author (daiyuchao@gmail.com).}
}
\affiliations{
    \textsuperscript{\rm 1}School of Electronics and Information, Northwestern Polytechnical University, and Shaanxi Key Laboratory of Information Acquisition and Processing, Xi'an, China, \\
    \textsuperscript{\rm 2}TapTap, \\
    \textsuperscript{\rm 3}OpenNLPLab
    
}

\usepackage{bibentry}

\begin{document}

\maketitle

\begin{abstract}

Vision Transformers (ViTs) have revolutionized computer vision, yet their self-attention mechanism lacks explicit spatial inductive biases, leading to suboptimal performance on spatially-structured tasks. Existing approaches introduce data-independent spatial decay based on fixed distance metrics, applying uniform attention weighting regardless of image content and limiting adaptability to diverse visual scenarios. Inspired by recent advances in large language models where content-aware gating mechanisms (e.g., GLA, HGRN2, FOX) significantly outperform static alternatives, we present the first successful adaptation of data-dependent spatial decay to 2D vision transformers. 
We introduce \textbf{Spatial Decay Transformer (SDT)}, featuring a novel Context-Aware Gating (CAG) mechanism that generates dynamic, data-dependent decay for patch interactions. 
Our approach learns to modulate spatial attention based on both content relevance and spatial proximity. We address the fundamental challenge of 1D-to-2D adaptation through a unified spatial-content fusion framework that integrates manhattan distance-based spatial priors with learned content representations.
Extensive experiments on ImageNet-1K classification and generation tasks demonstrate consistent improvements over strong baselines. Our work establishes data-dependent spatial decay as a new paradigm for enhancing spatial attention in vision transformers.

\end{abstract}


\input{1.intro}
\input{2.relatedworks}
\input{3.method}

\input{4.experiments}
\input{5_conclusion}

\section{Acknowledgments}
This research was supported in part by the National Natural Science Foundation of China (62271410, 12150007). Yuxin Mao is sponsored by the Innovation Foundation for Doctor Dissertation of NWPU (CX2024014).

\bibliography{aaai2026}

\end{document}

%% file: 1.intro.tex
\section{Introduction}

Vision Transformers (ViTs)~\cite{dosovitskiyimage_vit} have fundamentally transformed computer vision and multimodal tasks~\cite{zhou2022audio,zhou2024avss,zhou2024label,zhou2024vaplan,zhou2025mettle,zhou2025towards,zhou2025clasp,mao2024tavgbench,mao2023multimodal,mao2024generative}, achieving state-of-the-art performance across diverse tasks from image classification to generation~\cite{peebles_dit_iccv_2023}. 
The cornerstone of their success lies in the self-attention mechanism~\cite{vaswani2017attention}, which enables global receptive fields and long-range dependency modeling by treating images as sequences of patches. 
However, this design choice introduces a critical limitation: the inherent permutation equivariance of self-attention renders it agnostic to the 2D spatial structure of images, treating spatially adjacent patches identically to distant ones.

This spatial blindness poses significant challenges, as models must learn fundamental spatial relationships purely from data, often requiring extensive training to achieve satisfactory performance. 
To this end, recent works have sought to inject spatial inductive biases directly into the attention mechanism. 
Retentive Networks (RMT)~\cite{fan2024rmt} exemplify this approach by introducing data-independent spatial decay matrices that apply fixed, distance-based attention weighting according to spatial proximity. While providing useful locality bias, this strategy suffers from fundamental rigidity: the same spatial decay pattern is uniformly applied regardless of image content, preventing adaptive focus on semantically relevant regions.

Recent advances in large language models (LLMs)~\cite{li2025minimax,chen2025minimax} offer compelling insights that challenge this static approach. 
In linear attention mechanisms, data-dependent decay has emerged as a superior paradigm, with models like GLA~\cite{yanggated}, HGRN2~\cite{qinhgrn2,mao2025autoregressive}, and Mamba2~\cite{dao_mamba2_iclr} demonstrating that content-aware gating significantly outperforms data-independent counterparts. The Forgetting Transformer~\cite{linforgetting} validates this principle within standard quadratic attention, introducing learnable forget gates that dynamically modulate attention based on input content rather than fixed positional relationships. This data-dependent approach enables more nuanced, context-sensitive information flow, leading to substantial improvements in sequence modeling.

Motivated by these successes, we investigate \emph{whether data-dependent decay can be effectively adapted for 2D visual tasks}. 
This translation from 1D sequential to 2D spatial domains presents unique challenges: unlike sequential data with natural linear temporal relationships, images exhibit complex 2D spatial topologies requiring careful consideration of both horizontal and vertical interactions. Furthermore, extending from 1D positional decay to 2D spatial decay demands novel architectural designs that effectively capture content-dependent spatial relationships while maintaining computational efficiency.

We introduce \textbf{Spatial Decay Transformer}, pioneering the application of data-dependent spatial decay to vision transformers through a novel \textbf{Context-Aware Gating (CAG)} mechanism. Our approach generates dynamic, content-dependent decay strengths for 2D patch interactions, enabling selective modulation of spatial attention based on both content relevance and spatial proximity. 
Unlike the fixed manhattan distance-based decay of RMT, our method adapts to the semantic content of each image.

Our technical contribution addresses the 1D-to-2D adaptation challenge through a unified spatial-content fusion framework that integrates manhattan distance-based spatial priors with learned content representations. For computational efficiency in high-resolution stages, we propose a decomposed implementation that maintains content-dependent characteristics while reducing memory complexity.

Extensive experiments on ImageNet-1K classification and generation demonstrate consistent improvements over strong baselines including RMT, with particularly notable gains in tasks requiring fine-grained spatial understanding. Comprehensive ablation studies validate the superiority of data-dependent over data-independent spatial decay, confirming that content-aware gating is crucial for optimal 2D spatial attention.

Our contributions are threefold:

\begin{compactitem}
\item We systematically identify limitations of data-independent spatial decay in vision transformers and demonstrate the necessity of content-aware spatial attention mechanisms.
\item We introduce the first successful adaptation of data-dependent decay from 1D sequential modeling to 2D spatial attention, pioneering a new paradigm for spatial bias injection in vision transformers.
\item We propose Spatial Decay Transformer with Context-Aware Gating, achieving significant performance improvements and establishing a strong baseline for spatial attention in vision transformers.
\end{compactitem}

%% file: 2.relatedworks.tex
\section{Related Works}

\noindent\textbf{Vision Transformers and Spatial Biases.}
Vision Transformers (ViTs)~\cite{dosovitskiyimage_vit,qin2024lightnet} achieve superior performance over CNNs in image recognition but require massive datasets due to lacking spatial inductive biases. DeiT~\cite{deit} addresses this limitation through knowledge distillation from CNN teachers, enabling effective training on standard datasets like ImageNet-1K~\cite{krizhevsky2012imagenet}.
Subsequent research focuses on incorporating spatial information through various position encoding schemes, including absolute~\cite{dosovitskiyimage_vit}, relative~\cite{shaw2018self}, rotary~\cite{su2024roformer}, and conditional~\cite{chuconditional} position embeddings. Hierarchical transformers such as PVT~\cite{pvtv2}, Swin Transformer~\cite{SwinTransformer}, and MViT~\cite{li2022mvitv2} build multi-scale feature pyramids to improve dense prediction tasks. Recent hierarchical approaches include FasterViT~\cite{hatamizadehfastervit} with hierarchical attention for computational efficiency.

\noindent\textbf{Explicit Spatial Priors in Attention.}
To address the fundamental lack of spatial inductive biases in standard self-attention, researchers have explored directly incorporating spatial awareness into the attention mechanism. This approach typically involves modifying the attention matrix with explicit, data-independent spatial priors defined by geometric distances between patches.
Hybrid architectures like CoAtNet~\cite{dai2021coatnet} and BoTNet~\cite{srinivas2021bottleneck} effectively combine convolution and attention. Most relevant to our work, Retentive Vision Transformer (RMT)~\cite{fan2024rmt} introduces fixed spatial decay based on Manhattan distance, extending the 1d temporal decay mechanism of RetNet~\cite{sun2023retentive} to spatial domains through explicit spatial decay matrices.

\noindent\textbf{Data-Dependent Decay in Sequence Modeling.}
Recent breakthroughs in natural language processing and sequence modeling demonstrate the superiority of data-dependent state dynamics over static positional information. 
This paradigm shift is exemplified in the evolution of linear attention mechanisms~\cite{choromanski_performer_iclr_2020,katharopoulos_icml_transformerrnn_icml_2020} and state-space models (SSMs)~\cite{gu2023mamba}. Mamba~\cite{dao_mamba2_iclr} and related architectures utilize input-dependent state transitions, enabling selective information retention and forgetting based on current token content. 
This principle is generalized across architectures: Gated Linear Attention (GLA)~\cite{yanggated} and HGRN2~\cite{qinhgrn2} explicitly incorporate content-aware gating into linear recurrent models, achieving significant performance improvements over data-independent counterparts.
Particularly relevant is the Forgetting Transformer~\cite{linforgetting}, which integrates learnable forget gates into self-attention, demonstrating that content-based modulation outperforms fixed positional relationships for 1D sequences.

\noindent\textbf{Uniqueness of Our Solutions.}
Building upon these advances, our work pioneers the adaptation of data-dependent mechanisms to 2D spatial domains. While previous approaches focus on temporal decay patterns, we address the unique challenges of 2D topology, including bidirectional spatial dependencies and the need for principled spatial distance metrics that respect image geometry. 
To our knowledge, we present the first content-aware, dynamic spatial decay mechanism for vision transformers, bridging static spatial priors and dynamic data-dependent mechanisms.

%% file: 3.method.tex
\section{Method}

\subsection{Preliminaries}

\textbf{Vision Transformer Foundation.} Vision Transformers process images by partitioning them into non-overlapping patches, treating each patch as a token in a sequence. Given input features $\mathbf{X} \in \mathbb{R}^{L \times D}$ where $L = H \times W$ represents spatial resolution and $D$ denotes feature dimension, standard self-attention computes:

\begin{equation}
\mathbf{O} = \text{softmax}\left({\mathbf{Q}\mathbf{K}^T}/{\sqrt{d_k}}\right)\mathbf{V}
\end{equation}
where $\mathbf{Q}, \mathbf{K}, \mathbf{V} \in \mathbb{R}^{L \times d_k}$ are query, key, and value projections with linear projection matrices $\mathbf W_{q}, \mathbf W_{k}, \mathbf W_{v}$ respectively. While this mechanism enables global receptive fields, it lacks inherent spatial awareness, treating all patch interactions uniformly regardless of their geometric relationships.

\noindent\textbf{Data-Dependent Attention Modulation.} Recent advances in large language models have demonstrated the superiority of content-aware gating over fixed positional biases. In the autoregressive setting, this mechanism applies content-dependent decay where the output at position $i$ accumulates information with learned decay strengths:

\begin{equation}
\mathbf{o}_i = \sum_{j=1}^{i} \alpha_{ij} \mathbf{v}_j, \alpha_{ij} \propto \exp\left(\mathbf{q}_i^T \mathbf{k}_j + \sum_{l=j}^{i-1} \mathbf{g}_l\right)
\end{equation}
where $\mathbf{g}_l$ represents learned gating values controlling information decay. This recurrent formulation naturally captures the cumulative nature of content-dependent attention modulation.

For parallel computation, this mechanism can be expressed in matrix form. More generally, given input $\mathbf{X}$, we compute content-dependent modulation logits and apply:

\begin{equation}
\mathbf{O} = \text{Softmax}\left(\mathbf{S} + \mathbf{B}_{\text{mod}}\right)
\label{equ:base_atten}
\end{equation}
where $\mathbf{S} = {\mathbf{Q}\mathbf{K}^T}/{\sqrt{d_k}}$ represents the standard attention scores and $\mathbf{B}_{\text{mod}}$ is the content-dependent bias matrix derived from modulation logits. This framework provides a general representation for content-aware attention weighting, where different constructions of $\mathbf{B}_{\text{mod}}$ can encode various attention preferences from temporal dependencies in sequences to spatial relationships in images.

\subsection{Context-Aware Gating}

\noindent\textbf{Design Motivation.} 
Traditional spatial attention mechanisms employ fixed positional encodings~\cite{dosovitskiyimage_vit,shaw2018self} or distance-based decay~\cite{fan2024rmt} that apply uniform spatial biases regardless of image content. This approach fundamentally ignores semantic relationships: semantically related regions should maintain strong attention connections regardless of spatial distance, while irrelevant regions should be suppressed even when spatially adjacent. We propose Context-Aware Gating (CAG) to address this limitation through dynamic, content-dependent spatial attention modulation.

\begin{figure}[t]
  \centering
    \includegraphics[width=0.85\linewidth]{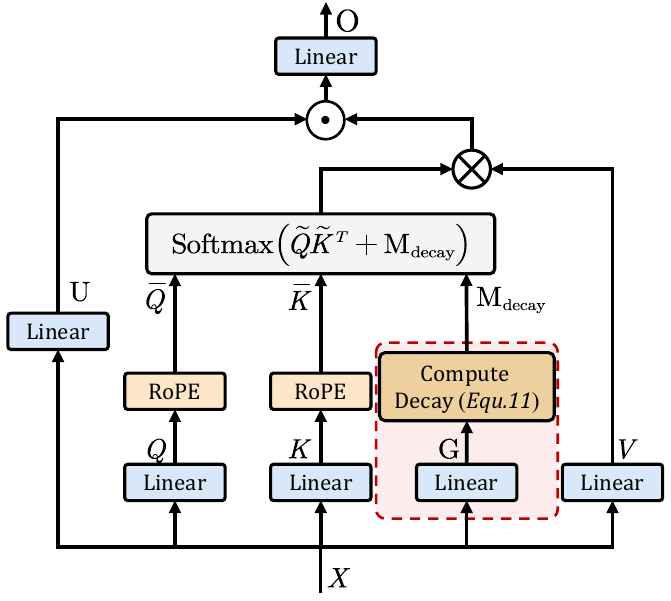}
    \caption{{The network structure of the Spatial Decay Layer}. The attention weights are modulated by a learned decay map $\mathbf{M}_{\text{decay}}$ computed from $\mathbf{G}$, enabling spatially adaptive attention.}
     \label{arch}
\end{figure}

\noindent\textbf{Content-Dependent Gate Generation.} 
Given input features $\mathbf{X} \in \mathbb{R}^{B \times H \times W \times D}$, we generate head-specific decay logits through learnable projection:

\begin{equation}
\mathbf{F} = \mathbf{X}\mathbf{W}_g \in \mathbb{R}^{B \times H \times W \times N},
\end{equation}
where $\mathbf{W}_g \in \mathbb{R}^{D \times N}$ is a learnable parameter matrix and $N$ is the number of attention heads. This projection enables the model to learn head-specific spatial attention patterns, allowing different heads to focus on different types of spatial relationships (e.g., local texture patterns vs. global object structures).

\noindent\textbf{Bounded Decay Computation.} The decay logits are transformed into bounded decay strengths through a log-sigmoid activation:

\begin{equation}
\mathbf{G} = \log\sigma(\mathbf{F}) \in \mathbb{R}^{B \times L \times N},
\end{equation}
where $L = H \times W$ and spatial dimensions are flattened for computational efficiency. The log-sigmoid transformation ensures decay strengths are bounded in $(-\infty, 0]$, providing stable gradient flow during training.


\begin{figure*}[t]
  \centering
    \includegraphics[width=0.85\linewidth]{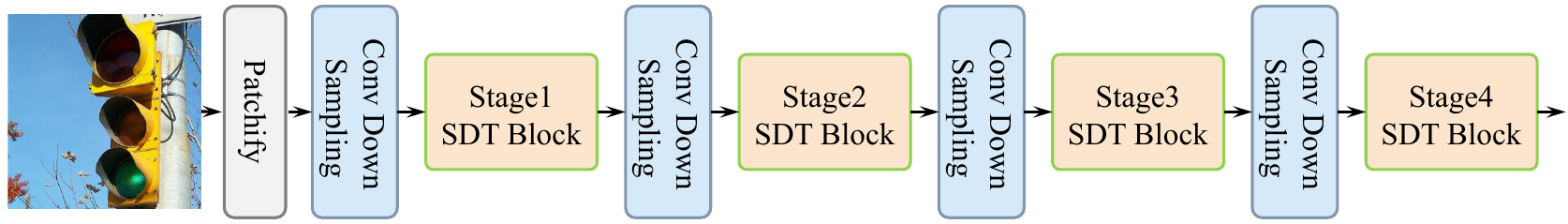}
    \caption{{Overall architecture} of the proposed Learnable Spatial Decay based Vision Transformer. The model consists of four stages of Spatial Decay Transformer (SDT) Blocks, and each stage consists of several Spatial Decay Layers as shown in Fig.~\ref{arch}.}
     \label{arch:all}
\end{figure*}

\subsection{Spatial Decay Extension: From 1D to 2D}

\noindent\textbf{Theoretical Challenges.} The extension of data-dependent decay from 1D sequential modeling to 2D spatial attention constitutes a non-trivial theoretical challenge. In 1D causal attention~\cite{yanggated,qinhgrn2}, temporal ordering enables efficient cumulative computation:

\begin{equation}
\mathbf{M}_{\text{1D}}[i,j] = \sum_{k=j}^{i-1} \mathbf{g}_k, \quad \text{for } i > j.
\label{equ:1d_decay}
\end{equation}

However, 2D spatial grids lack inherent ordering, presenting three challenges: 
(1) \textit{bidirectional dependencies} where each position interacts with neighbors in all directions, 
(2) \textit{non-causal relationships} where spatial proximity lacks temporal precedence, and 
(3) \textit{topological complexity} requiring sophisticated distance metrics beyond sequential indexing.

\noindent\textbf{Spatial Distance Formulation.} We establish a mathematical framework for spatial distance computation in discrete 2D grids. Given spatial positions $i$ and $j$ with coordinates $\mathbf{p}_i = (h_i, w_i)$ and $\mathbf{p}_j = (h_j, w_j)$ respectively, we define the manhattan distance metric:

\begin{equation}
d_{\text{M}}(\mathbf{p}_i, \mathbf{p}_j) = |h_i - h_j| + |w_i - w_j|.
\end{equation}

This choice is theoretically motivated by its natural alignment with discrete grid topology, computational efficiency, and proven effectiveness in spatial modeling tasks. 
The manhattan distance preserves the grid-aligned structure inherent in vision transformers while providing a principled measure of spatial proximity.



\noindent\textbf{Content-Dependent Spatial Fusion Framework.} 
We propose a unified spatial-content fusion mechanism that bridges fixed 2D image geometric priors and adaptive content representations. 

For notational clarity in the subsequent formulation, we omit the batch dimension $B$. Thus, we consider the gate for a single batch item as $\mathbf{G} \in \mathbb{R}^{L \times N}$.
The combined decay strength $\mathbf{M}_{\text{combined}}[i,j]$ (which is an $N$-dimensional vector) between any two spatial positions $i, j \in [1, \dots, L]$ is then defined as:
\begin{equation}
\mathbf{M}_{\text{combined}}[i,j] = \frac{1}{2}(\mathbf{G}[i,:] + \mathbf{G}[j,:]) \cdot d_{\text{M}}(\mathbf{p}_i, \mathbf{p}_j) \cdot \alpha,
\label{equ:decay_fusion}
\end{equation}
where $\mathbf{G}[i,:] \in \mathbb{R}^{N}$ and $\mathbf{G}[j,:] \in \mathbb{R}^{N}$ are the content-dependent gating vectors for spatial positions $i$ and $j$, respectively. 
This clarifies the ambiguity raised by the reviewer: we are combining the $N$-dimensional gating vectors corresponding to the $i$-th and $j$-th spatial locations.
The rationale for using their average, $\frac{1}{2}(\mathbf{G}[i,:] + \mathbf{G}[j,:])$, is to establish a symmetric and balanced measure of mutual influence. 
This operation ensures that the content-dependent modulation between position $i$ and position $j$ is reciprocal and jointly determined by the content at both locations. 
This averaged gating value then modulates the fixed scalar distance $d_{\text{M}}(\mathbf{p}_i, \mathbf{p}_j)$ (which is broadcast across all feature dimensions), and $\alpha \in \mathbb{R}^+$ is a scaling factor.






\noindent\textbf{Final Spatial Decay Mask and Integration.} The complete spatial decay mask is computed as:

\begin{equation}
\mathbf{M}_{\text{decay}}[i,j] = -|\mathbf{M}_{\text{combined}}[i,j]|.
\end{equation}
The negative absolute value operation serves dual purposes: 
(1) it maps all values to the non-positive domain, ensuring attention score reduction rather than amplification, and 
(2) it maintains gradient flow stability by preventing unbounded positive values during backpropagation. 
Based on Equ.~\eqref{equ:base_atten}, the final attention computation becomes:

\begin{equation}
\mathbf{O} = \text{Softmax}\left({\mathbf{Q}\mathbf{K}^T}/{\sqrt{d_k}} + \mathbf{M}_{\text{decay}}\right)\mathbf{V}.
\end{equation}

\noindent\textbf{Efficient Decomposed Implementation.} In hierarchical vision architectures, early stages with high spatial resolution face computational challenges when computing the full $L \times L$ spatial decay mask $\mathbf{M}_{\text{decay}}$, leading to excessive memory consumption. To address this challenge, we propose a decomposed implementation that is selectively applied only to the high-resolution stages.

The decomposed approach separately computes attention scores for horizontal and vertical directions, applying one-dimensional data-dependent decay to each direction:

\begin{equation}
\begin{gathered}
\mathbf{F}_H = \mathbf{X}\mathbf{W}_{g,H},
\mathbf{F}_W = \mathbf{X}\mathbf{W}_{g,W}, \\
\mathbf{M}_{\text{decay},H}[i,j] = -\sum_{k=\min(i_h,j_h)}^{\max(i_h,j_h)-1} |\log\sigma(\mathbf{F}_H[k])|, \\
\mathbf{M}_{\text{decay},W}[i,j] = -\sum_{k=\min(i_w,j_w)}^{\max(i_w,j_w)-1} |\log\sigma(\mathbf{F}_W[k])|.
\end{gathered}
\end{equation}

The decomposed attention is then computed as:

\begin{equation}
\mathbf{O} = \text{Attn}_H (\text{Attn}_W \mathbf{V})^T,
\end{equation}
where $\text{Attn}_H$ and $\text{Attn}_W$ apply the respective decay masks. This decomposition reduces complexity from $\mathcal{O}(L^2)$ to $\mathcal{O}(H^2 + W^2)$ while maintaining the data-dependent characteristics of our gating mechanism. For the latter two stages with reduced spatial resolution, we employ the full 2D spatial decay mask for optimal performance.

\begin{table*}[ht]
    \centering
    \setlength{\tabcolsep}{0.45mm}
    \scalebox{1}{
    \begin{tabular}{l|c c|c}
        \toprule[1pt]
        Model & \makecell{Parmas\\(M)} & \makecell{FLOPs\\(G)} & \makecell{Acc\\(\%)}\\
        \midrule[0.5pt]
        PVTv2-b1~\cite{pvtv2} & 13 & 2.1 & 78.7 \\
        QuadTree-B-b1~\cite{quadtree} & 14 & 2.3 & 80.0 \\
        RegionViT-T~\cite{regionvit} & 14 & 2.4 & 80.4 \\
        MPViT-XS~\cite{mpvit} & 11 & 2.9 & 80.9 \\
        tiny-MOAT-2~\cite{MOAT} & 10 & 2.3 & 81.0 \\
        VAN-B1~\cite{VAN} & 14 & 2.5 & 81.1 \\
        BiFormer-T~\cite{biformer} & 13 & 2.2 & 81.4 \\
        Conv2Former-N~\cite{conv2former} & 15 & 2.2 & 81.5 \\
        CrossFormer-T~\cite{crossformer} & 28 & 2.9 & 81.5 \\
        NAT-M~\cite{NAT} & 20 & 2.7 & 81.8 \\
        QnA-T~\cite{QnA} & 16 & 2.5 & 82.0 \\
        GC-ViT-XT~\cite{globalvit} & 20 & 2.6 & 82.0 \\
        SMT-T~\cite{SMT} & 12 & 2.4 & 82.2 \\
        RMT-T~\cite{fan2024rmt} & 14 & 2.5 & 82.4 \\
        SDT-H-T (Ours) & 14 & 2.7 & \textbf{82.7} \\
        \midrule[0.5pt]
        DeiT-S~\cite{deit} & 22 & 4.6 & 79.9 \\
        Swin-T~\cite{SwinTransformer} & 29 & 4.5 & 81.3 \\
        ConvNeXt-T~\cite{convnext} & 29 & 4.5 & 82.1 \\
        Focal-T~\cite{focal} & 29 & 4.9 & 82.2 \\
        FocalNet-T~\cite{focalnet} & 29 & 4.5 & 82.3 \\
        RegionViT-S~\cite{regionvit} & 31 & 5.3 & 82.6 \\
        CSWin-T~\cite{cswin} & 23 & 4.3 & 82.7 \\
        MPViT-S~\cite{mpvit} & 23 & 4.7 & 83.0 \\
        ScalableViT-S~\cite{ScalableViT} & 32 & 4.2 & 83.1\\        
        \bottomrule[1pt]
    \end{tabular}}
    \setlength{\tabcolsep}{0.45mm}
    \scalebox{1}{
    \begin{tabular}{l|c c|c}
        \toprule[1pt]
        Model & \makecell{Parmas\\(M)} & \makecell{FLOPs\\(G)} & \makecell{Acc\\(\%)}\\
        \midrule[0.5pt]
        MOAT-0~\cite{MOAT} & 28 & 5.7 & 83.3 \\
        Ortho-S~\cite{Ortho} & 24 & 4.5 & 83.4 \\
        CMT-S~\cite{cmt} & 25 & 4.0 & 83.5 \\
        MaxViT-T~\cite{maxvit} & 31 & 5.6 & 83.6 \\
        SMT-S~\cite{SMT} & 20 & 4.8 & 83.7 \\
        BiFormer-S~\cite{biformer} & 26 & 4.5 & 83.8 \\
        RMT-S~\cite{fan2024rmt} & 27 & 4.5 & 84.1 \\
        SDT-H-S (Ours) & 27 & 4.8 & \textbf{84.2} \\
        \bottomrule[1pt]
        Swin-S~\cite{SwinTransformer} & 50 & 8.7 & 83.0 \\
        ConvNeXt-S~\cite{convnext} & 50 & 8.7 & 83.1 \\
        CrossFormer-B~\cite{crossformer} & 52 & 9.2 & 83.4 \\
        NAT-S~\cite{NAT} & 51 & 7.8 & 83.7 \\
        Quadtree-B-b4~\cite{quadtree} & 64 & 11.5 & 84.0 \\
        Ortho-B~\cite{Ortho} & 50 & 8.6 & 84.0 \\
        ScaleViT-B~\cite{ScalableViT} & 81 & 8.6 & 84.1 \\
        MOAT-1~\cite{MOAT} & 42 & 9.1 & 84.2 \\
        InternImage-S~\cite{internimage} & 50 & 8.0 & 84.2 \\
        DaViT-S~\cite{davit} & 50 & 8.8 & 84.2 \\
        GC-ViT-S~\cite{globalvit} & 51 & 8.5 & 84.3 \\
        BiFormer-B~\cite{biformer} & 57 & 9.8 & 84.3 \\
        MViTv2-B~\cite{mvitv2} & 52 & 10.2 & 84.4 \\
        iFormer-B~\cite{iformer} & 48 & 9.4 & 84.6 \\
        RMT-B~\cite{fan2024rmt} &  54 &  9.7 &  85.0 \\
        SDT-H-B (Ours) & 54 & 10.8 & \textbf{85.1} \\
        \bottomrule[1pt]
    \end{tabular}}
    \caption{{Performance comparison for image classification task on ImageNet-1K.}}
    \label{tab:ImageNet}
    \vspace{-4mm}
\end{table*}

\subsection{Overall Architecture}
\noindent\textbf{Spatial Decay Attention Layer.}
The Spatial Decay Attention Layer is composed of a Spatial Decay Attention (SDA) and a Feed-Forward Network (FFN).
Each SDA block integrates: 
(1) Multi-head Context-Aware Spatial Gating that generates head-specific decay logits, 
(2) Rotary Position Embedding~\cite{su2024roformer} for enhanced positional awareness, 
(3) Local Position Encoding~\cite{chuconditional} through depthwise convolutions, and 
(4) A low-rank output gate for comprehensive attention control, as shown in Fig.~\ref{arch}.
The architecture seamlessly combines hierarchical feature learning with adaptive, content-aware spatial attention for superior performance across diverse vision tasks.

\begin{equation}
\mathbf{O} = \mathbf{U} \odot \left[\text{Softmax}\left(\frac{\tilde{\mathbf{Q}}\tilde{\mathbf{K}}^T}{\sqrt{d_k}} + \mathbf{M}_\text{decay}\right)\mathbf{V} + \text{LPE}(\mathbf{V})\right],
\end{equation}
where $\tilde{\mathbf{Q}} = R(\mathbf{Q})$ and $\tilde{\mathbf{K}} = R(\mathbf{K})$ are RoPE-enhanced queries and keys.
$\mathbf{Q},\mathbf{K},\mathbf{V}$ is projected via linear projection matrices $\mathbf{W}_q, \mathbf{W}_k, \mathbf{W}_v$.
$\mathbf{U} = \sigma(\mathbf{X}\mathbf{W}_{u_1})\mathbf{W}_{u_2}$ with $\mathbf{W}_{u_1}, \mathbf{W}_{u_2}$ is low-rank projection matrices for output gating, and $\text{LPE}(\cdot)$ denotes local position encoding via depthwise convolutions.

\noindent\textbf{Hierarchical Design.} 
Our Spatial Decay Transformer (SDT) can be implemented in both hierarchical architecture and plain structure configurations. 
When employing the hierarchical architecture (SDT-H), the model adopts a multi-stage design using the downsampling strategy~\cite{SwinTransformer}, with structural design aligned with that of RMT. 
As shown in Fig.~\ref{arch:all}, SDT-H comprises four stages, each progressively reducing spatial resolution while increasing feature dimension. 
This hierarchical design enables effective multi-scale feature learning, which is crucial for vision tasks. 
We employ a decomposed implementation in the first two stages and a global implementation in the final two stages. 
Alternatively, SDT can be configured as a plain structure (SDT-P) without hierarchical downsampling for scenarios requiring consistent spatial resolution throughout the network.

%% file: 4.experiments.tex
\begin{table}[t]
    \setlength{\tabcolsep}{0.10cm}
    \begin{tabular}{lcccccc}
    \toprule
    Model & FID$\downarrow$   & sFID$\downarrow$  & IS$\uparrow$     & Pre.$\uparrow$ & Rec.$\uparrow$ & Params \\
    \midrule
    ADM   & 10.94 & 6.02 & 100.98 & 0.69 & 0.63 & - \\
    ADM-U & 7.49 & 5.13 & 127.49 & 0.72 & 0.63 & - \\
    ADM-G & 4.59 & 5.25 & 186.70 & 0.82 & 0.52 & - \\
    \midrule
    CDM & 4.88 & - & 158.71 & - & - &- \\
    \midrule
    LDM-8 & 15.51 & - & 79.03 & 0.65 & 0.63 &395M \\
    LDM-8-G & 7.76 & - & 209.52 & {0.84} & 0.35  &506M \\
    LDM-4 & 10.56 & - & 103.49 & 0.71 & 0.62  &400M \\
    LDM-4-G & 3.60 & - & 247.67 & \textbf{0.87} & 0.48  &400M  \\
    \midrule
    DiT-S  & 68.40 & -    & -      & -    & -             & 32M  \\
    DiT-B  & 43.47 & -    & -      & -    & -             & 130M \\
    DiT-L  & 23.33 & -    & -      & -    & -             & 459M \\
    DiT-XL & 19.47 & -    & -      & -    & -             & 675M \\
    DiT-XL & 9.62  & 6.85 & 121.50 & 0.67 & {0.67} & 675M \\
    DiT-XL-G & {2.27} & {4.60} & {278.24} & 0.83 & 0.57 &  675M\\
    \midrule
    SDT-P-S    & 60.70  & 10.42 & 23.15 & 0.40  & 0.60 & 33M \\
    SDT-P-B    & 37.47  & 6.82  & 60.49 & 0.53  & 0.60 & 132M \\
    SDT-P-L    & 21.28  & 5.99  & 63.79 & 0.62 & 0.64  & 462M \\
    SDT-P-XL   & 18.20  & 5.74  & 71.82 & 0.64 & 0.63  & 679M \\
    SDT-P-XL   & 6.82   & 6.21  & 158.34& 0.71 & 0.64  & 679M \\
    SDT-P-XL-G & \textbf{2.25}   & \textbf{4.59}  & \textbf{279.85}& {0.83} & \textbf{0.58}    & 679M \\
    \bottomrule
    \end{tabular}
    \caption{{Performance comparison for image generation task on ImageNet-1K.} 
    SDT-P-XL achieves state-of-the-art FID with or without classifier-free guidance (-G).
    \enquote{Pre.} and \enquote{Rec.} represents the \enquote{Precision} and \enquote{Recall}.
    The best result is highlighted with \textbf{bold}.}
    \label{table:image_generation}
\end{table}

\section{Experiments}
We evaluate the performance and scalability of our proposed Spatial Decay Transformer (SDT) as a substitute for existing models on image classification (using SDT-H) and image generation (using SDT-P) tasks.

\subsection{Settings}
\noindent\textbf{Image Classification.}
We train all models from scratch on the ImageNet-1K dataset~\cite{krizhevsky2012imagenet}. 
For fair comparison, we adopt the same training protocol as in~\cite{fan2024rmt}, relying solely on classification loss as supervision. 
The stochastic depth rates for different size Tiny/Small/Base are set to 0.1, 0.15, and 0.4, respectively. 
We employ the AdamW~\cite{loshchilov2017decoupled_adamw} optimizer in conjunction with a cosine learning rate schedule. 
The initial learning rate is set to 0.001, with a weight decay of 0.05 and a batch size of 1024. 
For data augmentation and regularization, we follow the strong strategy used in~\cite{fan2024rmt}, including RandAugment~\cite{cubuk2020randaugment}, Mixup~\cite{zhang2018mixup}, CutMix~\cite{yun2019cutmix}, and Random Erasing~\cite{zhong2020random}.

\noindent\textbf{Image Generation.}
We build our model upon the Diffusion Transformer (DiT)~\citep{peebles_dit_iccv_2023}, employing our proposed SDT as the denoising network (Note that we are using a plain structure instead of a hierarchical structure).
We scale the model across various configurations (S, B, L, XL) with a fixed patch size of 2, consistent with DiT.
We conduct experiments on the ImageNet dataset~\citep{krizhevsky2012imagenet} at $256 \times 256$ resolution.
We compare performance against representative image generation methods, including ADM~\citep{dhariwal2021diffusion}, CDM~\citep{ho_CDM_JMLR_2022}, LDM~\citep{rombach_ldm_cvpr_2022}, and DiT~\citep{peebles_dit_iccv_2023}.
All models are trained for 400K steps with a batch size of 256 to evaluate scaling capabilities.
For the largest model variant, we extend training to 0.8M steps with a batch size of 512 (compared to 7M steps in DiT) to enhance generative performance.

\begin{figure}[ht]
  \centering
    \includegraphics[width=0.95\linewidth]{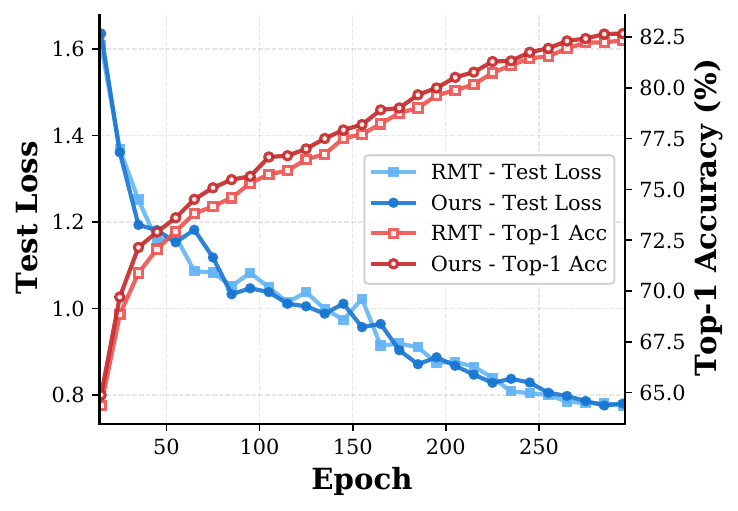}
    \caption{{Training loss comparison between our proposed SDT-H-T and RMT-T}. 
    The blue curve represents our SDT-H-T, and the orange curve represents RMT-T.}
    \vspace{-4mm}
     \label{fig:loss}
\end{figure}

\subsection{Main Results}

\noindent\textbf{Image Classification.}
We evaluate our models against state-of-the-art vision transformers on the ImageNet-1K validation set. As shown in Table~\ref{tab:ImageNet}, our approach demonstrates competitive performance across various model scales.

Our method consistently outperforms the data-independent spatial decay method, RMT. This comparison provides strong evidence for the effectiveness of data-dependent, context-aware gating over static, content-agnostic decay mechanisms. Additionally, our models achieve competitive results with other leading architectures. 
Notably, SDT-S surpasses established models including ConvNeXt-S, BiFormer-S, and NAT-S, demonstrating strong performance within its complexity class. 
The consistent improvements across different scales validate the robustness and scalability of our spatial decay mechanism.

Fig.~\ref{fig:loss} presents training dynamics comparing our SDT-H-T with RMT-T. The results show that our model achieves faster convergence and superior final accuracy, further supporting the effectiveness of the proposed context-aware spatial decay approach.



\noindent\textbf{Image Generation.}
We evaluate this variant, termed SDT-P, on class-conditional image generation using ImageNet at $256 \times 256$ resolution, as shown in Table~\ref{table:image_generation}. 
We can observe that SDT-P-XL-G achieves competitive performance with a FID of 2.25, marginally outperforming DiT-XL-G (2.27 FID) while maintaining comparable parameter count. 
These results suggest that the enhanced spatial reasoning capabilities of our model translate effectively to generative tasks. Specifically, the ability to dynamically attend to relevant contextual information appears beneficial for image generation.
The improved performance indicates that our Context-Aware Gating (CAG) mechanism provides beneficial guidance during the denoising process, contributing to more coherent output generation.

\begin{figure}[t]
  \centering
    \includegraphics[width=0.95\linewidth]{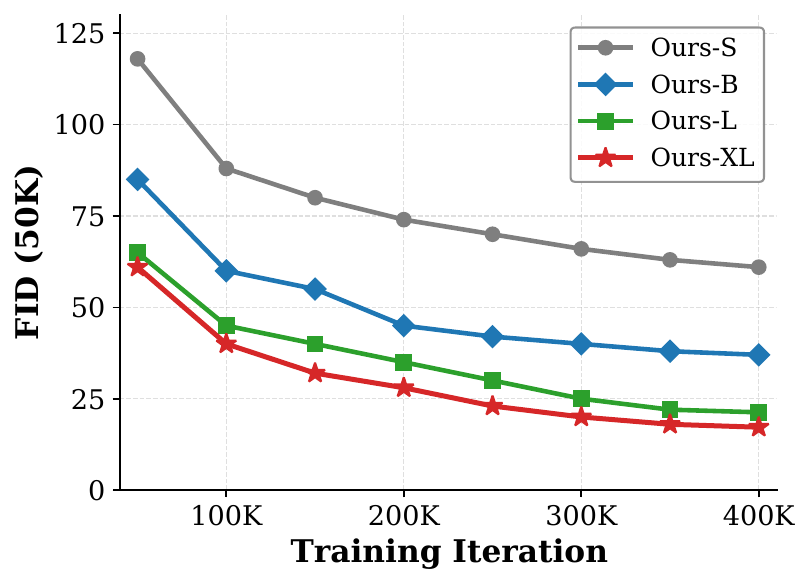}
    \caption{{Scaling up the SDT-P enhances the FID during every iterations of training.} 
    We present the FID-50K across training iterations for four SDT-P models. Enhancing the SDT-P backbone results in improved generative models for all sizes of models.}
     \label{fig:fid}
\end{figure}

We further analyze the scaling properties of our approach across four model sizes on ImageNet. As illustrated in Figure~\ref{fig:fid}, performance consistently improves from S to XL scale, demonstrating favorable scaling characteristics. This trend suggests the potential applicability of our method to larger-scale diffusion models, where spatial attention mechanisms become increasingly important for maintaining generation quality at higher resolutions.




\subsection{Ablation Studies}
To dissect our model and validate the contributions of its key components, we conduct a series of ablation studies. 
We extend our analysis beyond image classification and image generation to demonstrate the robustness and general applicability of our design principles. 
All experiments use the Tiny (Ours-T) configuration for classification (reporting Top-1 Acc) and a comparable Base (Ours-B) model for generation (reporting FID).

\noindent\textbf{Effectiveness of the Context-Aware Gating mechanism.}
We propose a Context-Aware Gating mechanism that enables data-dependent decay computation to learn the importance of correlations between image patches. Unlike traditional fixed decay patterns, this mechanism adaptively determines decay values based on the actual content and context of the input data, providing more flexible and effective attention modeling.
To thoroughly evaluate the effectiveness of our Context-Aware Gating mechanism, we conduct ablation studies with the following variants:
(1) No Decay: We remove all decay mechanisms and employ only the original attention mechanism without any positional or content-based decay modulation. This serves as our baseline to demonstrate the necessity of decay mechanisms.
(2) Spatial Decay: We implement a data-independent decay approach that computes decay values solely based on the relative positional relationships between image patches. This variant captures spatial proximity but ignores content-dependent correlations.
(3) Context-Aware Gating (Ours): Our proposed data-dependent decay mechanism that learns to adaptively weight patch correlations based on both spatial relationships and content similarity.

The quantitative comparison is presented in Table~\ref{tab:ablation_decay}. The results demonstrate the superiority of our Context-Aware Gating mechanism. The baseline without decay shows limited performance, highlighting the importance of incorporating decay mechanisms. The Spatial Decay variant provides improvements by considering positional relationships, but its data-independent nature limits its adaptability. In contrast, our Context-Aware Gating achieves the best performance by dynamically adjusting decay patterns based on input content, effectively capturing both spatial and semantic correlations between patches.

\begin{table}[t]
\centering
\setlength{\tabcolsep}{0.2cm}
\begin{tabular}{lcc}
\toprule
{Decay Mechanism} & {Acc (\%)$\uparrow$} & {FID$\downarrow$} \\
\midrule
No decay      & 81.9 & 43.22 \\
Spatial Decay & 82.4 & 40.29 \\
\midrule
\textbf{Ours} & \textbf{82.7} & \textbf{37.47} \\
\bottomrule
\end{tabular}
\caption{{Ablation study on Context-Aware Gating variants} on image classification (Acc) and generation (FID).}
\label{tab:ablation_decay}
\end{table}

\begin{table}[t]
\centering
\setlength{\tabcolsep}{0.2cm}
\begin{tabular}{lcc}
\toprule
{Decay Mechanism} & {Acc (\%)$\uparrow$} & {FID$\downarrow$} \\
\midrule
1D Decay            & 82.2 & 42.25 \\
Bidirectional Decay & 82.3 & 41.32 \\
Decomposed 2D Decay & 82.5 & 39.11 \\
\midrule
\textbf{Ours} & \textbf{82.7} & \textbf{37.47} \\
\bottomrule
\end{tabular}
\caption{{Ablation study on Content-Dependent Spatial Fusion variants} on image classification (Acc) and generation (FID).}
\label{tab:ablation_fusion}
\end{table}

\noindent\textbf{Effectiveness of the Content-Dependent Spatial Fusion.}
In extending the 1D decay mechanism to 2D scenarios, we propose a Content-Dependent Spatial Fusion Framework (CDSF) that integrates spatial relationships with learnable decay patterns. This framework combines the manhattan distance between image patches with computed learnable decay values for spatially-aware attention.
To validate the effectiveness of our spatial fusion mechanism, we design several variants:
(1) 1D Decay: We treat the 2D image as a flattened 1D sequence and directly apply the decay computation from Equ.~(\ref{equ:1d_decay}).
(2) Bidirectional Decay: Based on 1D decay, we adopt a bidirectional scanning strategy inspired by VisionMamba~\cite{zhu2024visionmamba}, averaging the forward and backward decay computations.
(3) Decomposed 2D Decay: We decompose the image along height and width dimensions, computing decay values independently for each dimension.
(4) Ours Content-Dependent Spatial Fusion: Our proposed framework that integrates both spatial distance and content-dependent decay computation.

As shown in Table~\ref{tab:ablation_fusion}, our proposed CDSF consistently outperforms all variants. 
The 1D decay shows limited performance due to ignoring 2D spatial structure. 
Bidirectional Decay provides modest improvements through bidirectional processing. 
The decomposed approach captures dimension-specific patterns but lacks cross-dimensional interaction. 
In contrast, our CDSF achieves superior performance by effectively combining spatial proximity with content-dependent decay, validating our design.

\begin{table}[t]
\centering
\setlength{\tabcolsep}{0.2cm}
\begin{tabular}{lcc}
\toprule
{Decay Mechanism} & {Acc (\%)$\uparrow$} & {FID$\downarrow$} \\
\midrule
$\alpha=0.05$          & 82.5 & 40.57 \\
$\mathbf{\alpha=0.1}$  & \textbf{82.7} & \textbf{37.47} \\
$\alpha=0.15$          & 82.6 & 39.45 \\
$\alpha=0.2$           & 82.4 & 39.79 \\
\bottomrule
\end{tabular}
\caption{\textbf{Ablation study on the $\mathbf{\alpha}$ factor} in Equ.~(\ref{equ:decay_fusion}) on image classification (Acc) and generation (FID).}
\label{tab:ablation_alpha}
\end{table}

\noindent\textbf{Impact of $\mathbf{\alpha}$.}
As shown in Equ.~\ref{equ:decay_fusion}, alpha controls the strength of decay, larger values indicate stronger decay. We conduct experiments to verify the impact of alpha. The experimental results, shown in Table~(\ref{tab:ablation_alpha}), show that an alpha of 0.1 achieves optimal decay strength.

%% file: 5_conclusion.tex
\section{Conclusion}
We have introduced Spatial Decay Transformer, which successfully adapts data-dependent decay mechanisms from language models to 2D spatial attention in vision transformers. Our Context-Aware Gating (CAG) mechanism generates dynamic, content-dependent spatial decay that adapts to image content, overcoming the rigidity of fixed distance-based approaches like RMT.
Through a unified spatial-content fusion framework, we address the fundamental challenge of extending 1D sequential decay to 2D spatial domains. Extensive experiments on ImageNet-1K classification and generation demonstrate consistent improvements over strong baselines, validating that content-aware spatial gating is crucial for optimal 2D attention.
Our work establishes data-dependent spatial decay as a new paradigm for vision transformers, opening promising directions for other spatial-structured vision tasks or spatial-temporal vision tasks.